%% file: conference_101719.tex
\documentclass[conference]{IEEEtran}
\IEEEoverridecommandlockouts
\usepackage{cite}

\usepackage{amsmath,amssymb,amsfonts}

\usepackage{array}
\usepackage{booktabs}
\usepackage{multirow}
\usepackage{tabularx}

\usepackage{graphicx}
\usepackage{subcaption}

\usepackage{xcolor}
\usepackage{soul}
\definecolor{lightyellow}{RGB}{255,255,204}
\sethlcolor{lightyellow}

\usepackage{textcomp}
\usepackage{url}
\usepackage[justification=centering]{caption}
\usepackage{xspace}

\usepackage{algorithm}
\usepackage{algorithmicx}
\usepackage{algpseudocode}

\usepackage{stfloats}

\def\BibTeX{{\rm B\kern-.05em{\sc i\kern-.025em b}\kern-.08em
    T\kern-.1667em\lower.7ex\hbox{E}\kern-.125emX}}

\begin{document}

\title{\vspace{0.20in}PRISA: Proactive Infrastructure LiDAR Framework for Intersection Safety Assessment \\
\thanks{This research was partially supported by the Federal Highway Administration under Agreement No. 693JJ32350028. It was also partially supported by the Transportation Network Growth Opportunity (TNGO) initiative, funded by the Tennessee Department of Economic and Community Development in collaboration with the University of Tennessee at Chattanooga and industry partners. Any opinions, findings, and conclusions or recommendations expressed in this publication are those of the authors and do not necessarily reflect the view of the Federal Highway Administration.  \\
\copyright~2026 IEEE. Personal use of this material is permitted.
Permission from IEEE must be obtained for all other uses, in any current
or future media, including reprinting/republishing this material for
advertising or promotional purposes, creating new collective works,
for resale or redistribution to servers or lists, or reuse of any
copyrighted component of this work in other works.}}

\author{
\IEEEauthorblockN{
Tam Bang\IEEEauthorrefmark{1},
Hussam Abubakr\IEEEauthorrefmark{1},
Emiliano de la Garza Villarreal\IEEEauthorrefmark{1},
Truc Phuong Nguyen\IEEEauthorrefmark{1},
Austin Harris\IEEEauthorrefmark{1}
}
\IEEEauthorblockN{
Toru Hirano\IEEEauthorrefmark{2},
Mina Sartipi\IEEEauthorrefmark{1},
Yunfei Xu\IEEEauthorrefmark{2},
Hoang H. Nguyen\IEEEauthorrefmark{1}
}
\IEEEauthorblockA{\IEEEauthorrefmark{1}Center for Urban Informatics and Progress, University of Tennessee at Chattanooga, USA}
\IEEEauthorblockA{\IEEEauthorrefmark{2}DENSO International America, Inc., Southfield, MI 48033, USA}
}

\maketitle

\input{sec/abstract}
\input{sec/introduction}
\input{sec/related_work}
\input{sec/methodology}
\input{sec/experiment}
\input{sec/ablation}
\input{sec/conclusion}

{
    \small
    \bibliographystyle{IEEEtran}
    \bibliography{ref}
}

\end{document}

%% file: sec/abstract.tex
\begin{abstract}

Urban intersections are among the most hazardous locations in road networks, posing significant risks to vehicles and vulnerable road users (VRUs) such as pedestrians and cyclists. The complexity of multi-agent interactions demands continuous, real-time monitoring systems capable of anticipating conflicts before they escalate into crashes. We present PRISA, a modular infrastructure LiDAR framework leveraging privacy-preserving, low-light-robust roadside sensors for long-term traffic observation and real-time risk detection at the edge. The framework comprises two core components: a sensing and perception layer and a plug-and-play risk assessment module. The latter automatically curates site-specific training data from accumulated perception outputs to train a trajectory prediction model without manual annotation. It then deploys the trained model for continuous motion forecasting and dual surrogate safety evaluation, using Time-to-Collision (TTC) for longitudinal conflicts and Predicted Post-Encroachment Time (PPET) for crossing and VRU-involved interactions. PRISA is evaluated on the public R-LiViT dataset and deployed on an NVIDIA Jetson AGX Thor at a live signalized intersection in Chattanooga, Tennessee. PPET-based assessment operates at 194~ms end-to-end latency over a 2.4-second predictive horizon, with TTC-based detection and perception remaining within real-time constraints, demonstrating practical feasibility for proactive multi-agent intersection safety monitoring.
\end{abstract}


%% file: sec/introduction.tex
\section{Introduction}
Urban intersections are among the most hazardous locations in road networks, accounting for a significant share of traffic fatalities and severe injuries. The increasing presence of vulnerable road users (e.g., pedestrians, cyclists) further compounds the safety challenge, as these groups are highly susceptible to injury in conflict events~\cite{world2024global}. Infrastructure-mounted proactive safety monitoring systems capable of detecting potential conflicts in real time and issuing early warnings are therefore critical for evidence-based intervention. Such systems must be deployable with minimal technical expertise across diverse roadside environments.

Roadside sensors offer a persistent, wide-area perspective that is well suited for intersection monitoring. LiDAR is well suited for roadside perception due to its robustness in low-light environments and its precise 3D spatial measurements, enabling accurate localization, heading, and velocity estimation independent of illumination~\cite{zhang2026roadside}. Critically, LiDAR captures no visually identifiable information, making it privacy-preserving.

Proactive conflict detection requires not only observing current road user states, but anticipating future interactions. Surrogate Safety Measures (SSMs) --- quantitative metrics of near-miss severity widely adopted as surrogate indicators of crash risk~\cite{arun2021systematic} --- form the foundation of such systems. Predictive SSMs, such as PPET~\cite{qi2020modified}, computed from forecast trajectories provide earlier and more reliable warnings than reactive metrics alone. However, deploying this pipeline in practice introduces two key challenges. First, learning-based prediction models require site-specific trajectory data. However, raw perception outputs often contain noisy, fragmented, and incomplete tracks that degrade model performance. This necessitates automated, self-supervised data curation from accumulated observations to generate reliable training samples without manual annotation. Second, safety-critical real-time operation demands deterministic execution, which cloud offloading cannot guarantee due to variable network delays, making on-device edge deployment a practical necessity.

Existing work addresses these challenges in isolation. Many methods are designed solely for offline analysis of pre-recorded datasets~\cite{nikolaou2023review}, while others face challenges in bridging upstream perception outputs to deployable conflict detection, limiting their generalization across heterogeneous intersection environments~\cite{kumar2025surrogate}. Furthermore, prior work often adopts vehicle-centric evaluations, whereas safe urban intersections demand a system-wide infrastructure perspective capturing complex multi-agent interactions across all road users.

To overcome these limitations, we present PRISA (\underline{PR}oactive Infrastructure LiDAR Framework for \underline{I}ntersection \underline{S}afety \underline{A}ssessment), a real-time, edge-deployable, modular framework for intersection safety monitoring using infrastructure LiDAR. Trajectory prediction and conflict evaluation are centralized into a cohesive, plug-and-play risk assessment module, fully decoupled from the upstream perception layer and compatible with any compliant detection and tracking system. The key contributions of PRISA are as follows:

\begin{itemize}

    
    

    \item \textbf{Modular, edge-deployable safety framework:} An end-to-end pipeline that transforms raw roadside LiDAR data into actionable conflict metrics, operating in real time on embedded edge hardware without cloud dependency.
    
    \item \textbf{Plug-and-play risk assessment module:} A self-contained module integrating self-supervised trajectory prediction and dual-SSM conflict evaluation, operating on a standardized state vector and independent of the upstream perception system.

    \item \textbf{Real-world validation:} The proposed pipeline is validated on a public roadside dataset and further deployed on an NVIDIA Jetson AGX Thor edge device at a live urban intersection in Chattanooga, Tennessee, demonstrating its real-time operational feasibility.

\end{itemize}

%% file: sec/related_work.tex
\section{Related Work}

\subsection{Surrogate Safety Measures and Conflict Assessment}
Surrogate Safety Measures offer a lightweight yet effective proactive alternative to crash-based evaluation, quantifying the severity of traffic conflicts and near-misses without requiring actual crash event data. The FHWA Surrogate Safety Assessment Model (SSAM) established the foundational methodology, defining algorithms for computing TTC for rear-end conflicts and Post-Encroachment Time (PET) for crossing conflicts from simulated vehicle trajectories~\cite{gettman2003surrogate, gettman2008surrogate}. Subsequent validation studies confirmed that simulation-derived conflict metrics correlate significantly with field-measured crash data at signalized intersections~\cite{essa2013vissim}. Beyond simulation, both measures have been extended to field deployments at unsignalized intersections under mixed traffic conditions, where TTC is most appropriate for longitudinal rear-end interactions and PET is best suited for lateral crossing events involving diverging or merging paths~\cite{singh2024conflict, arun2023review}.

Reactive SSMs are, however, inherently constrained to observed states, limiting the intervention lead time available to safety systems. Predicted Post-Encroachment Time extends the classical PET concept to forecast trajectories, enabling proactive rather than reactive conflict detection~\cite{li2024realtime}. Adopting a dual-measure strategy --- TTC for longitudinal following conflicts and PPET for crossing and VRU-involved interactions --- therefore provides both broader conflict coverage and earlier warning capability than either metric alone~\cite{pascucci2023before}. Despite the maturity of learning-based trajectory prediction methods with demonstrated accuracy, such as MID~\cite{gu2022stochastic} and Trajectron++~\cite{salzmann2020trajectron++}, their integration into infrastructure-side risk assessment pipelines has remained limited. Raw perception outputs from roadside sensors are inherently noisy, fragmented, and incomplete, requiring careful data curation before they can serve as reliable training inputs, while the substantial computational resources required for real-time roadside deployment further constrain practical adoption~\cite{huang2025trajectory}.

\subsection{Infrastructure-Based Risk Assessment Frameworks}
Intersection safety monitoring has been approached across a range of sensing modalities and computational paradigms, with the shared goal of translating raw sensor data into actionable conflict indicators. Camera and UAV-based systems have been widely adopted for SSM computation, leveraging accessible deployment and broad spatial coverage: UAV platforms enable accurate bird's-eye trajectory extraction for PET-based behavioral conflict analysis at both signalized and unsignalized locations~\cite{chen2017surrogate, lamsal2025post}, while fixed roadside camera systems support real-time computation of TTC and PET directly from live video streams~\cite{jalayer2022real}. Learning-based methods have advanced conflict detection further by applying deep learning to vehicle trajectory data at signalized intersections, moving beyond purely reactive metrics toward anticipatory conflict prediction~\cite{zhang2024real}. Infrastructure LiDAR offers precise 3D spatial measurement independent of lighting conditions and has been deployed for intersection monitoring across several contexts, ranging from work-zone safety and VRU conflict detection~\cite{nrel2024lidar, spie2025lidar} to the integration of trajectory prediction with SSM-based risk assessment using roadside-mounted sensors~\cite{xu2024roadside}.

Despite this progress, existing frameworks face several challenges. Camera-based systems remain constrained by perspective distortion and limited depth accuracy, while learning-based approaches typically depend on large-scale manually annotated datasets and cloud-level computational resources, limiting their feasibility for edge deployment. More fundamentally, perception and risk assessment are often treated as isolated concerns: SSM-based systems assume clean trajectory inputs derived from simulation or manual annotation, rather than raw outputs from an operational perception pipeline. To the best of our knowledge, no prior work has integrated a self-supervised training pipeline with a modular risk assessment module capable of interfacing directly with heterogeneous upstream perception layers while operating in real time on embedded edge hardware at a live urban intersection. The proposed framework directly addresses these gaps.

%% file: sec/methodology.tex
\section{Methodology}

\begin{figure}[h]
\centering
\captionsetup{justification=justified, singlelinecheck=false}
\includegraphics[width=0.49\textwidth]{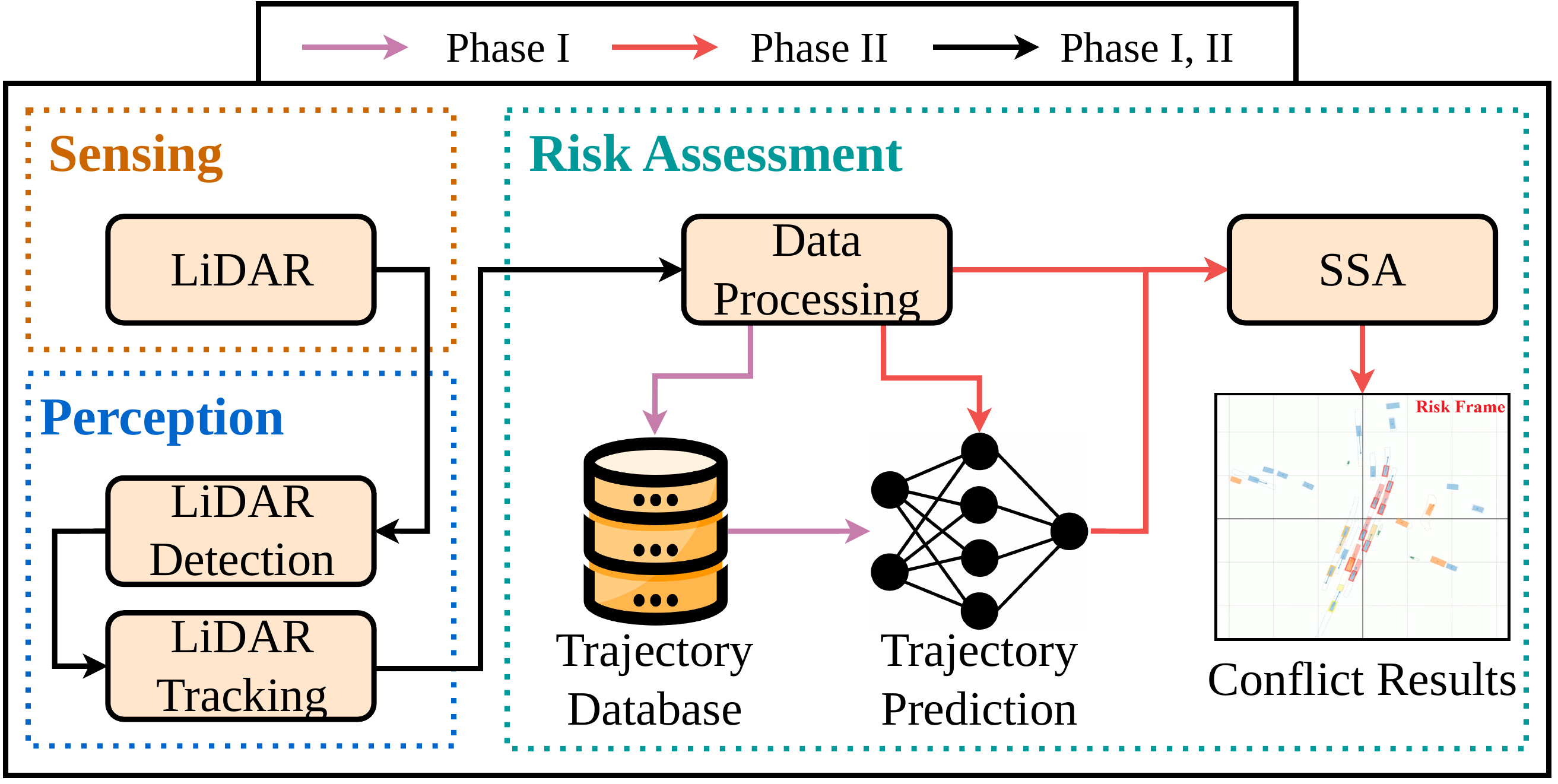}
\caption{Overview of PRISA, comprising the upstream Sensing and Perception layer and the Plug-and-Play Risk Assessment module with its two operational phases.}
\label{fig:pipeline}
\end{figure}

PRISA comprises two components, as illustrated in Fig.~\ref{fig:pipeline}. The first is the \textit{sensing and perception} layer, which processes raw LiDAR point clouds into standardized tracked object states. The second is the \textit{plug-and-play risk assessment module}, operating in two sequential phases. In Phase~I, site-specific training data are automatically curated from accumulated perception outputs to train a trajectory prediction model without manual annotation. In Phase~II, the trained model runs continuous real-time inference, with predicted trajectories fed directly into surrogate safety evaluation using TTC and PPET. Requiring only a standardized state vector as input, the module is fully decoupled from the upstream perception layer and compatible with any compliant detection and tracking system.

\subsection{LiDAR Sensing and Perception}
The sensing layer uses LiDAR sensors to provide continuous 3D point cloud coverage of the intersection area. The perception module applies 3D object detection and multi-object tracking directly on the point cloud data to identify and track road users, extracting bounding box parameters (center position, dimensions, heading) and semantic class labels $c \in \{\text{Vehicle}, \text{VRU}\}$, where VRU encompasses pedestrians, cyclists, and other vulnerable road users. While the perception operates in full 3D, downstream modules consume only the bird's-eye-view (BEV) coordinates $(x, y)$. This two-dimensional representation is sufficient for road-level conflict analysis, where longitudinal and lateral positions are the primary factors governing interactions, and substantially reduces computational overhead.

For each tracked object $i$ at frame $t$, the perception module produces a state vector
\begin{equation}
    \mathbf{s}_i^t = \left(x_i^t,\; y_i^t,\; v_{x,i}^t,\; v_{y,i}^t,\;
                     \theta_i^t,\; l_i,\; w_i,\; c_i\right),
\end{equation}
where $(x_i^t, y_i^t)$ denotes the 2D center position in the global coordinate frame, $\mathbf{v}_i^t = (v_{x,i}^t, v_{y,i}^t)$ is the 2D velocity vector, $\theta_i^t$ is the heading angle, $(l_i, w_i)$ are the length and width of the bounding box, and $c_i$ is the semantic class. The scalar speed is defined as $v_i^t = \|\mathbf{v}_i^t\|_2$. The set of all tracked states $\mathcal{S}^t = \{\mathbf{s}_i^t\}$ serves as input to the plug-and-play risk assessment module.
\subsection{Plug-and-Play Risk Assessment Module}
\subsubsection{Phase I: Self-Supervised Data Curation and Model Training}
In Phase~I, the perception output $\mathcal{S}^t$ is processed through a 
data curation pipeline (Algorithm~\ref{alg:data-processing}) that 
accumulates object trajectories over extended observation periods. The 
pipeline filters out incomplete or noisy tracks, interpolates brief gaps, 
and downsamples stationary objects to create a clean trajectory database. 
This automated curation eliminates the need for manual ground-truth labeling: 
the large volume of accumulated data, combined with track quality filtering, 
provides sufficient supervision for training the prediction model in a 
self-supervised manner.

A learning-based prediction model is trained on this curated database to forecast $T_\text{pred}$ future positions from $T_\text{obs}$ historical positions. Phase~I terminates when the model converges, after which the trained weights are frozen and deployed in Phase~II.

\subsubsection{Phase II: Real-Time Trajectory Prediction}
To reduce computational overhead, a speed-based filtering step is applied to vehicles before trajectory prediction. The average speed of object $i$ over the observation window, $\bar{v}_i$, is then used to partition objects into three categories:

\begin{itemize} 
    
    \item \textbf{Dynamic vehicles} ($c_i = \text{Vehicle}, \bar{v}_i > v_\text{thr}$): The prediction model forecasts the future trajectory $\hat{\mathcal{T}}_i$, and these objects proceed to risk assessment. 
    
    \item \textbf{Slow / Stationary vehicles} ($c_i = \text{Vehicle}, \bar{v}_i \leq v_\text{thr}$): Prediction is skipped to save computation. For risk assessment, their current state is held constant ($\hat{\mathbf{s}}_i^{t+k} = \mathbf{s}_i^t$), treating them as static obstacles (e.g., queued or parked cars) for collision checking. 

    \item \textbf{VRUs}: Trajectory prediction is always applied regardless of speed, as VRU motion is inherently less predictable and their trajectories are essential for PPET calculation.

    \end{itemize}

This class-dependent filtering strategy balances computational efficiency with safety: trajectory prediction — the most computationally expensive step — is applied selectively for vehicles based on motion, while all VRUs receive full trajectory forecasting to capture their potentially erratic movement patterns.

\begin{algorithm}[h]
\small
\caption{Data Processing for Trajectory Prediction}
\label{alg:data-processing}
\textbf{Require:} Perception outputs $\mathcal{S}^t$, observation window $T_\text{obs}$, prediction horizon $T_\text{pred}$ \\
\textbf{Phase I (Training):}
\begin{algorithmic}[1]
\State Accumulate frames over $T_\text{DB}$ to build trajectory database $\mathcal{D}$
\State Group detections by track ID to obtain per-object sequences $\{\tau^i\}$
\State Filter: Discard tracks with length $< T_\text{obs} + T_\text{pred}$ or gaps $> T_\text{obs}$
\State Interpolate remaining gaps using linear interpolation
\State Downsample stationary tracks (standing duration $> \tau_\text{stand}$) to reduce redundancy
\State Train prediction model on processed dataset until convergence
\end{algorithmic}
\textbf{Phase II (Inference):}
\begin{algorithmic}[1]
\State Maintain sliding window of $T_\text{obs}$ frames for each active track
\State Filter: Remove tracks with $< 2$ consecutive detections
\State Interpolate brief gaps ($\leq 1$ frame) in active tracks
\State Extract observation sequence $(x_{t-T_\text{obs}+1}, \ldots, x_t)$ for each track
\State Apply prediction model to generate the future trajectory $\hat{\mathcal{T}}_i = \{(\hat{x}_i^{t+k},\hat{y}_i^{t+k})\}_{k=1}^{T_\text{pred}}$
\end{algorithmic}
\end{algorithm}

\subsubsection{Surrogate Safety Assessment (SSA)}

The Surrogate Safety Assessment (SSA) block, as the final stage of the risk assessment module, evaluates pairwise conflict risk among dynamic objects using two complementary SSMs. Object pairs are first classified by interaction type, the appropriate metric is applied, and the result is compared against a configurable threshold to determine conflict status. 

\paragraph{Interaction Classification}

For each object $i$, potential conflict partners are identified within a
velocity-dependent region of interest (ROI):
\begin{equation}
\small
    \text{ROI}_i = \left\{(x, y) : \|(x - x_i^t,\; y - y_i^t)\|_2 
    \leq R_\text{max} + \bar{v}_i \cdot T_\text{pred}\right\},
\end{equation}
where $(x, y)$ is any candidate position in the BEV plane, $(x_i^t, y_i^t)$ is the current position of object $i$, $R_\text{max}$ is a base detection radius, and $\bar{v}_i \cdot T_\text{pred}$ extends the search range proportionally to object speed.

For each pair $(i, j)$ within mutual ROIs, interaction type is determined by combining heading alignment and spatial position. The heading difference
\begin{equation}
    \Delta\theta_{ij} = \min\left(|\theta_i - \theta_j|,\;
                        360^\circ - |\theta_i - \theta_j|\right),
    \label{eq:heading_diff}
\end{equation}
measures angular alignment. To separate true following scenarios from 
side-by-side motion, the relative displacement $\mathbf{p}_j - \mathbf{p}_i$ 
is decomposed into longitudinal and lateral components with respect to $i$'s 
heading:
\begin{align}
    d_\text{long}^{ij} &= (\mathbf{p}_j - \mathbf{p}_i) \cdot
                          \begin{pmatrix}\cos\theta_i \\ \sin\theta_i\end{pmatrix}, \\
    d_\text{lat}^{ij}  &= (\mathbf{p}_j - \mathbf{p}_i) \cdot
                          \begin{pmatrix}-\sin\theta_i \\ \cos\theta_i\end{pmatrix}.
\end{align}

We define an adaptive corridor width $w_\text{corr} = \max\left((w_i + w_j)/2,\, w_\text{lane}\right)$, where $w_\text{lane} = 3.66$\,m, ensuring a physically meaningful minimum based on standard lane width \cite{global2016global}. Objects are then classified into three categories based on thresholds $\alpha_\text{long}$ (e.g., 30°) and $\alpha_\text{cross}$ (e.g., 60°):

\begin{itemize} 
    \item \textbf{Longitudinal (Following)}: Vehicle pairs where $\Delta\theta_{ij} < \alpha_\text{long}$, $d_\text{long}^{ij} > 0$ (leader is ahead), and $|d_\text{lat}^{ij}| < w_\text{corr}$. Evaluated using TTC.
    
    \item \textbf{Crossing / Lane-change}: All other vehicle pairs, including oblique merging ($\alpha_\text{long} \leq \Delta\theta_{ij} \leq \alpha_\text{cross}$) and perpendicular crossings ($\Delta\theta_{ij} > \alpha_\text{cross}$). Evaluated using PPET. 
    
    \item \textbf{VRU-involved}: Interactions involving at least one VRU, evaluated using PPET irrespective of heading alignment. Pedestrian–pedestrian pairs are excluded from conflict assessment. Vehicle–VRU pairs are also excluded when the vehicle is stationary or near-stationary (speed $\leq v_\text{thr}$), as stopped vehicles (e.g., buses at stops or parked cars) do not represent a genuine collision risk despite geometric proximity.
    
\end{itemize}

A single object may participate in multiple interactions simultaneously. Each pairwise interaction is evaluated independently, and the conflict with the highest severity (lowest TTC or PPET value) is prioritized for alert generation. All classification thresholds ($\alpha_\text{long}$, $\alpha_\text{cross}$, $\tau_\text{TTC}$, $\tau_\text{PPET}$) are configurable at runtime, enabling plug-and-play deployment across different intersection geometries.

\paragraph{Time-to-Collision}
For a longitudinal vehicle-to-vehicle pair (follower $f$, leader $\ell$), TTC is computed by accounting for vehicle dimensions rather than using centroids. The gap distance $d_{f\ell}$ is measured from the front-center of the follower to the rear-center of the leader, and the closing speed $\Delta v_{f\ell} = |v_f - v_\ell|$ is the difference in scalar velocities. TTC is then defined as:

\begin{equation}
    \text{TTC} =
    \begin{cases}
        \dfrac{d_{f\ell}}{\Delta v_{f\ell}} & \text{if } \Delta v_{f\ell} > 0,\\[6pt]
        \infty & \text{otherwise,}
    \end{cases}
    \label{eq:ttc}
\end{equation}
where a finite TTC indicates that the follower is approaching the leader; a conflict is flagged when $\text{TTC} < \tau_\text{TTC}$.

Each dynamic object is compared to its immediate leader — the nearest object ahead along its heading direction within the velocity-dependent ROI — avoiding redundant pairwise comparisons.

\paragraph{Predicted Post-Encroachment Time}
For crossing and lateral interactions, we adopt PPET, which extends the classical PET concept to forecast trajectories by propagating each object's predicted bounding boxes forward in time and identifying where and when their spatial extents intersect. Given a non-empty conflict zone $\mathcal{C}_{ij}$, the entry and exit times of each object are identified from the predicted swept volumes, and PPET is defined as:

\begin{equation}
    \text{PPET} =
    \begin{cases}
        t_\text{entry}^j - t_\text{exit}^i &
            \text{if } t_\text{exit}^i < t_\text{entry}^j,\\[4pt]
        t_\text{entry}^i - t_\text{exit}^j &
            \text{if } t_\text{exit}^j < t_\text{entry}^i,\\[4pt]
        0 & \text{otherwise (temporal overlap).}
    \end{cases}
    \label{eq:ppet}
\end{equation}

Algorithm~\ref{alg:ppet} formalizes the procedure for swept volume generation, dynamic conflict zone detection, and PPET evaluation based on Eq.~\ref{eq:ppet}. A conflict is declared when $\text{PPET} < \tau_\text{PPET}$.

\begin{algorithm}[h]
\small
\caption{PPET Computation for a Crossing Object Pair}
\label{alg:ppet}
\textbf{Require:} Predicted trajectories $\hat{\mathcal{T}}_i$,
$\hat{\mathcal{T}}_j$; prediction horizon $T_\text{pred}$ \\
\textbf{Output:} PPET value for pair $(i, j)$ at time $t$
\begin{algorithmic}[1]

\For{each object $o \in \{i, j\}$}
    \For{$k = 1$ to $T_\text{pred}$}
        \State Generate time-stamped bounding box $B_o^{t+k}$
    \EndFor
\EndFor

\State $\mathcal{C}_{ij} \gets \{ k \mid B_i^{t+k} \cap B_j^{t+k} \neq \emptyset,\ k \in \{1, \ldots, T_\text{pred}\} \}$

\If{$\mathcal{C}_{ij} = \emptyset$}
    \State \Return $\infty$ \quad \textit{(pair declared safe)}
\EndIf

\State Identify $t_\text{entry}^i, t_\text{exit}^i$ and
       $t_\text{entry}^j, t_\text{exit}^j$ from $\mathcal{C}_{ij}$
\If{neither object clears $\mathcal{C}_{ij}$ within $T_\text{pred}$}
    \State \Return $0$ \quad \textit{(unresolved conflict, treated as critical)}
\EndIf
\State Compute PPET via Eq.~\ref{eq:ppet}
\If{multiple conflict zones exist}
    \State \Return $\min(\text{PPET across all zones})$
\EndIf
\State \Return PPET
\end{algorithmic}
\end{algorithm}

%% file: sec/experiment.tex
\section{Experiments and Field Deployment}

The proposed framework is evaluated across two complementary settings. In the first, the risk assessment module is assessed in isolation on a public roadside LiDAR dataset using ground-truth trajectory inputs, enabling direct comparison against prior SSM-based methods on clean data. In the second, the complete end-to-end pipeline is deployed at a live urban intersection to validate real-world operational feasibility. All experiments are conducted on an NVIDIA Jetson AGX Thor operating in maximum performance mode to evaluate computational efficiency under target edge hardware conditions, and the perception module is treated as an interchangeable component throughout, demonstrating the plug-and-play design of the risk assessment module.

\subsection{Datasets and Experimental Setup}

\begin{figure*}[t]
\centering
\captionsetup{justification=justified, singlelinecheck=false}
\includegraphics[width=0.98\textwidth]{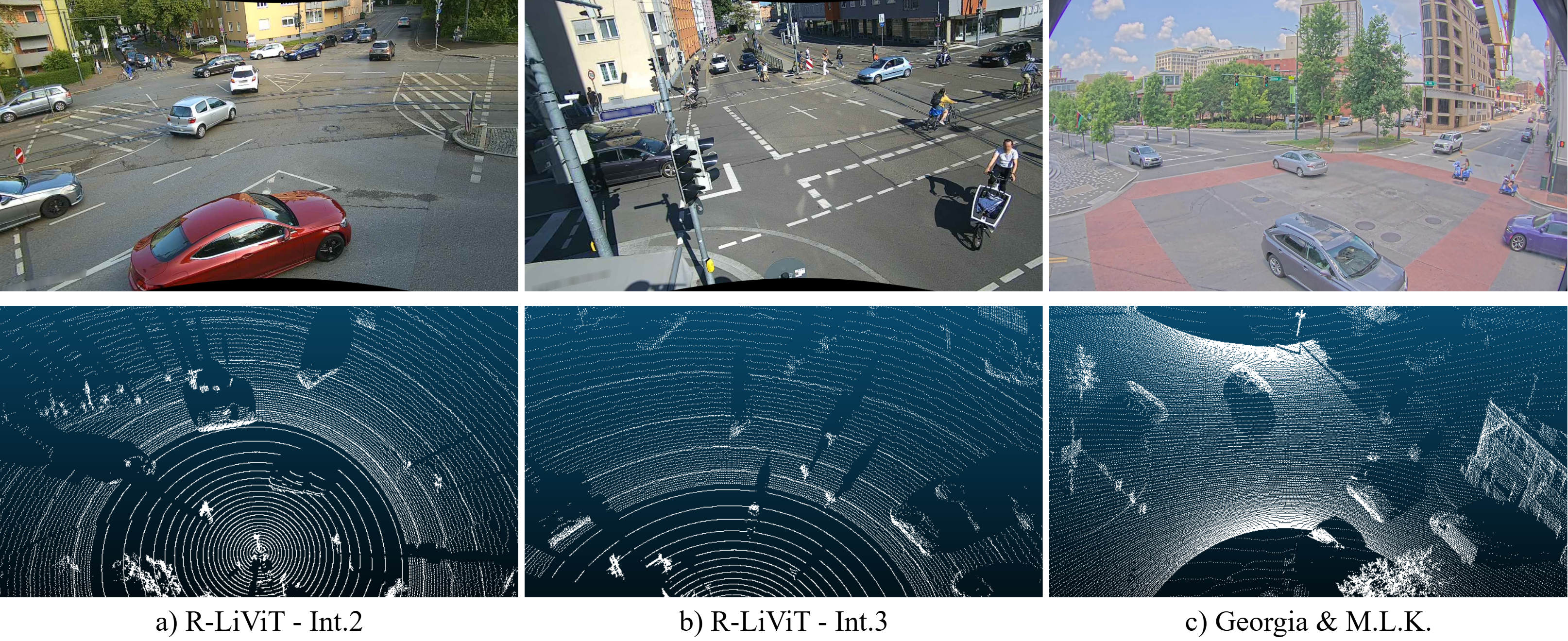}
\caption{RGB camera (top) and LiDAR point cloud (bottom) samples from one representative location within R-LiViT Int.2 and Int.3, and from the Georgia \& M.L.K. deployment site.}
\label{fig:dataset}
\end{figure*}

\subsubsection{R-LiViT Dataset}

The R-LiViT dataset~\cite{mirlach2025r} is a publicly available roadside LiDAR benchmark featuring high-density mixed traffic, including VRUs (pedestrian, cyclist, motorcycle) and vehicles (car, truck, bus, tramway), collected at three intersections. We utilize Intersection~2 (Locations~3, 4, 5) and Intersection~3 (Locations~6, 7), as both represent signalized urban settings with dense interactions. For each intersection, sequences are sorted chronologically per location and split 70\%/30\% by order: the first 70\% are used for trajectory prediction training and the last 30\% for risk assessment evaluation. Trajectory prediction metrics (ADE/FDE) are averaged across locations within each intersection, while conflict counts are summed across locations to reflect the total number of conflict events observed.

The dataset’s fragmented structure, consisting of short sequences of 50–200 frames collected across multiple separated sub-locations, is structurally incompatible with Phase I self-supervised training, which requires continuous long-duration streams from a fixed sensor deployment. We therefore use ground-truth trajectory annotations directly as inputs to the risk assessment module, consistent with established SSM evaluation practice~\cite{gettman2003surrogate, jalayer2022real}, isolating the dual-SSM contribution and enabling fair comparison against prior methods that operate on clean trajectory inputs.

\subsubsection{Field Deployment Site}
The full pipeline is deployed at the Georgia Avenue and M.L.K. Boulevard intersection in Chattanooga, Tennessee, a four-leg signalized intersection with mixed vehicle and VRU traffic. Two Ouster LiDAR sensors are mounted at a height of 5 meters at opposite corners, with point clouds unified by the Ouster BlueCity backend to provide 360$^\circ$ coverage~\cite{ousterbluecity}, supporting subclassification of vehicles (car, truck, bus, trailer) and VRUs (pedestrian, bicycle). Although the BlueCity API delivers perception outputs at 10~Hz, data are downsampled to 5~Hz to reduce inference frequency on the edge device and maintain a consistent frame interval across all experiments. Perception outputs are received via API and passed directly to the plug-and-play risk assessment module for downstream data curation, trajectory prediction, and conflict assessment. A total of 30 minutes of perception data are collected without manual annotation: 20 minutes for Phase~I self-supervised training and 10 minutes for Phase~II evaluation, with the two sessions recorded on separate days to ensure traffic pattern variability across training and evaluation.

Figure~\ref{fig:dataset} illustrates representative sensor views from both datasets, and Table~\ref{tab:dataset_stats} summarizes their key statistics.

\begin{table}[h]
\captionsetup{justification=justified, singlelinecheck=false}
\centering
\setlength{\tabcolsep}{0.5pt}
\renewcommand{\arraystretch}{1}
\scriptsize
\caption{Dataset overview at 5\,Hz.}
\label{tab:dataset_stats}
\begin{tabularx}{\columnwidth}{@{} l *{6}{>{\centering\arraybackslash}X} @{}}
\toprule
\multirow{3}{*}{\textbf{Metric}} 
& \multicolumn{4}{c}{\textbf{R-LiViT}} 
& \multicolumn{2}{c}{\textbf{Field}} \\
\cmidrule(lr){2-5} \cmidrule(lr){6-7}
& \multicolumn{2}{c}{\textbf{Int. 2}} 
& \multicolumn{2}{c}{\textbf{Int. 3}} 
& \multicolumn{2}{c}{\textbf{Georgia \& M.L.K.}} \\
\cmidrule(lr){2-3} \cmidrule(lr){4-5} \cmidrule(lr){6-7}
& \textbf{Train} & \textbf{Eval} 
& \textbf{Train} & \textbf{Eval} 
& \textbf{Train} & \textbf{Eval} \\
\midrule
Sequences      & 28   & 13   & 35   & 15   & --   & --   \\
Duration (min) & 6.8  & 3.5  & 8.3  & 4.0  & 20   & 10 \\
Frames         & 2050 & 1050 & 2500 & 1200 & 6000 & 3000 \\
Vehicle tracks & 772  & 207  & 873  & 123  & 1046 & 842  \\
VRU tracks     & 607  & 222  & 886  & 142  & 445  & 106  \\
\bottomrule
\end{tabularx}
\end{table}
\subsection{Trajectory Prediction Performance}
FlowChain~\cite{maeda2023fast}, MID~\cite{gu2022stochastic}, and
Trajectron++~\cite{salzmann2020trajectron++} are trained on all road user types jointly using $T_\text{obs} = 8$ frames and $T_\text{pred} = 12$ frames with a 70/15/15 train/validation/test split following~\cite{maeda2023fast}. Training data are prepared using the automated curation pipeline described in Algorithm~\ref{alg:data-processing}. All experiments adopt a uniform 5\,Hz frame rate ($\Delta t = 0.2$\,s), yielding 1.6\,s of observation history and a 2.4\,s prediction horizon, applied consistently across both the R-LiViT dataset and the downsampled field deployment data. Models are evaluated using Average Displacement Error (ADE) and Final Displacement Error (FDE).

Table~\ref{tab:traj_pred_all} presents prediction performance across all sites. Two kinematic baselines are included for reference: Constant Velocity (CV)~\cite{scholler2020constant}, which extrapolates the last observed velocity in a straight line, and Constant Acceleration (CA)~\cite{scholler2020constant}, which additionally accounts for the most recent acceleration. CV achieves competitive ADE/FDE across all sites due to the straight-path-dominant nature of intersection traffic. However, both baselines degrade substantially for turning maneuvers, which constitute the primary conflict scenarios at intersections. Furthermore, both methods produce a single deterministic trajectory, making them incompatible with the distribution-aware PPET metric used in risk assessment. Among the learned models, FlowChain achieves the lowest ADE and FDE across all sites, outperforming both MID and Trajectron++ by a substantial margin. Prediction time is measured for an average of 40 objects per frame with data already in the required input format, excluding preprocessing. Based on accuracy and computational efficiency, FlowChain is selected as the trajectory prediction backbone for all downstream risk assessment experiments.

\begin{table}[h]
\scriptsize
\captionsetup{justification=justified, singlelinecheck=false}
\centering
\setlength{\tabcolsep}{4pt}
\renewcommand{\arraystretch}{1}
\caption{Trajectory prediction ADE / FDE (metres) at 5\,Hz.}
\label{tab:traj_pred_all}
\begin{tabular}{lcccc}
\toprule
\multirow{2}{*}{\textbf{Model}}
& \multicolumn{2}{c}{\textbf{R-LiViT}}
& \textbf{Field}
& \multirow{2}{*}{\textbf{Pred. Time}} \\
\cmidrule(lr){2-3} \cmidrule(lr){4-4}
& \textbf{Int. 2} & \textbf{Int. 3} & \textbf{Georgia \& M.L.K.} & (ms)\\
\midrule
CV           & 2.432 / 4.656  & 2.239 / 4.167  & 1.796 / 3.625  & \textbf{0.05} \\
CA           & 5.001 / 12.707 & 5.262 / 13.422 & 4.410 / 11.282 & 0.11 \\
\midrule
MID          & 2.638 / 4.740  & 2.847 / 5.180  & 3.683 / 6.641  & 230 \\
Trajectron++ & 1.085 / 2.007  & 1.173 / 2.189  & 0.920 / 1.744  & 22  \\
FlowChain    & \textbf{0.646 / 1.111} & \textbf{0.491 / 0.710}
             & \textbf{0.505 / 0.928} & 28 \\
\bottomrule
\end{tabular}
\end{table}

\subsection{Risk Assessment Threshold Configuration}
Risk assessment thresholds are selected based on established traffic safety literature~\cite{gettman2008surrogate, yang2021proactive}. To identify appropriate operating thresholds, we perform a sensitivity analysis sweeping $\tau_\text{TTC}$ from 0.5\,s to 4.0\,s and $\tau_\text{PPET}$ from 0.5\,s to 2.4\,s, where the upper bound of the PPET sweep is governed by both our model's maximum prediction horizon and the FHWA guideline that PET\,$>$\,2.5\,s indicates a conflict-free interaction~\cite{gettman2003surrogate}. Conflict type is classified by approach angle following the geometry-based conflict taxonomy of Laureshyn et al.~\cite{laureshyn2010evaluation}: pairs with $\alpha < \alpha_\text{long} = 30^\circ$ are treated as following conflicts and pairs with $\alpha > \alpha_\text{cross} = 60^\circ$ as crossing conflicts. Near-stationary detections are suppressed using a speed threshold of $v_\text{thr} = 2.0$\,m/s and an ROI of 15.0\,m. For each threshold, we report total conflict instances and conflict rate (total conflict instances\,/\,total evaluation frames) across all sites in Table~\ref{tab:conflict_rates}.

As shown in Table~\ref{tab:conflict_rates}, at $\tau_\text{TTC} = 1.5$\,s the conflict rate remains below 100\% across all sites, indicating fewer than one simultaneous TTC conflict per frame on average, consistent with the standard TTC conflict criterion adopted in SSAM~\cite{gettman2003surrogate} and recent surrogate safety literature~\cite{lin2025predicted}. Beyond 2.0\,s, rates exceed 100\% at Int.2 and Int.3 (e.g., 102.7\% and 113.3\% at 2.5\,s), reflecting overlapping conflict windows from simultaneous multi-pair interactions rather than distinct critical events.

For PPET, the number of detected conflicts is already largely saturated at low thresholds, with only modest increases from 0.5\,s to 2.4\,s across all sites. Very small PPET values may arise not only from genuinely critical interactions, but also from limitations of finite-horizon prediction. First, if a conflict is detected near the end of the prediction horizon and the object does not leave the conflict zone within that horizon, the interaction is assigned PPET = 0 by definition. Second, the predictor may temporarily extend an approaching vehicle straight through the intersection or produce a turning trajectory that overlaps with stopped vehicles or nearby roadside objects, thereby yielding artificially low PPET values despite limited actual risk. We therefore select $\tau_{\text{PPET}} = 1.5$\,s as a practical operating threshold, as it captures most detected conflicts while providing a more stable operating point than very small thresholds, which are more sensitive to near-zero PPET cases, and remains well within the model's 2.4\,s prediction horizon.

\begin{table}[h]
\captionsetup{justification=justified, singlelinecheck=false}
\centering
\setlength{\tabcolsep}{0.5pt}
\renewcommand{\arraystretch}{1}
\scriptsize
\caption{Conflict detection sensitivity across thresholds (30\% evaluation split). Bold rows mark the selected thresholds ($\tau_\text{TTC} = \tau_\text{PPET} = 1.5$\,s).}
\label{tab:conflict_rates}
\begin{tabularx}{\columnwidth}{@{} c *{6}{>{\centering\arraybackslash}X} @{}}
\toprule
\multirow{2}{*}{\textbf{Thresh. (s)}}
& \multicolumn{4}{c}{\textbf{R-LiViT}}
& \multicolumn{2}{c}{\textbf{Field}} \\
\cmidrule(lr){2-5} \cmidrule(lr){6-7}
& \multicolumn{2}{c}{\textbf{Int. 2}}
& \multicolumn{2}{c}{\textbf{Int. 3}}
& \multicolumn{2}{c}{\textbf{Georgia \& M.L.K.}} \\
\cmidrule(lr){2-3} \cmidrule(lr){4-5} \cmidrule(lr){6-7}
& \textbf{Conflicts} & \textbf{Rate (\%)}
& \textbf{Conflicts} & \textbf{Rate (\%)}
& \textbf{Conflicts} & \textbf{Rate (\%)} \\
\midrule
\multicolumn{7}{c}{\textit{Time-to-Collision (TTC)}} \\
\midrule
0.5  & 192  & 18.3  & 168  & 14.0  & 169  & 5.6  \\
1.0  & 406  & 38.7  & 520  & 43.3  & 753  & 25.1 \\
\textbf{1.5} & \textbf{597} & \textbf{56.9} & \textbf{800} & \textbf{66.7} & \textbf{1190} & \textbf{39.7} \\
2.0  & 756  & 72.0  & 1041 & 86.8  & 1553 & 51.8 \\
2.5  & 915  & 87.1  & 1232 & 102.7 & 1944 & 64.8 \\
3.0  & 1082 & 103.0 & 1359 & 113.3 & 2320 & 77.3 \\
3.5  & 1232 & 117.3 & 1479 & 123.3 & 2595 & 86.5 \\
4.0  & 1353 & 128.9 & 1575 & 131.3 & 2861 & 95.4 \\
\midrule
\multicolumn{7}{c}{\textit{Predicted Post-Encroachment Time (PPET)}} \\
\midrule
0.5  & 597 & 56.9 & 729 & 60.8 & 1125 & 37.5 \\
1.0  & 612 & 58.3 & 753 & 62.8 & 1133 & 37.8 \\
\textbf{1.5} & \textbf{631} & \textbf{60.1} & \textbf{774} & \textbf{64.5} & \textbf{1145} & \textbf{38.2} \\
2.0  & 636 & 60.6 & 782 & 65.2 & 1152 & 38.4 \\
2.4  & 639 & 60.9 & 784 & 65.3 & 1152 & 38.4 \\
\bottomrule
\end{tabularx}
\end{table}

\subsection{Conflict Type Analysis}
To better understand the nature of detected conflicts, Table~\ref{tab:conflict_types} breaks down conflict pairs and frames by interaction type at the selected thresholds ($\tau_\text{TTC} = \tau_\text{PPET} = 1.5$\,s), distinguishing Vehicle--Vehicle, Vehicle--VRU, and VRU--VRU interactions across all sites.

Vehicle–Vehicle (TTC) conflicts dominate across all sites, with 284, 307, and 257 distinct pairs at Int.2, Int.3, and Georgia \& M.L.K., respectively. At the R-LiViT intersections, VRU-involved conflicts constitute a substantial portion of PPET detections (Int.2: 246 of 464 pairs and Int.3: 387 of 483 pairs), reflecting dense mixed-traffic conditions and the high VRU presence in the evaluation sets (222 and 142 VRU tracks).

Similarly, the elevated VRU–VRU counts at R-LiViT (78 pairs at Int.2 and 212 at Int.3) require careful interpretation. Many of these cases involve co-moving agents, such as a cyclist traveling alongside a pedestrian, whose trajectories satisfy the geometric PPET threshold without representing genuinely safety-critical interactions.

\begin{table}[h]
\captionsetup{justification=justified, singlelinecheck=false}
\centering
\setlength{\tabcolsep}{0.5pt}
\renewcommand{\arraystretch}{1}
\scriptsize
\caption{Conflict counts by interaction type at the selected thresholds ($\tau_\text{TTC} = \tau_\text{PPET} = 1.5$\,s). \textbf{Pairs} denotes unique object-pair combinations flagged as conflicting and \textbf{Frames} denotes total conflict instances across all frames.}
\label{tab:conflict_types}
\begin{tabularx}{\columnwidth}{@{} l *{6}{>{\centering\arraybackslash}X} @{}}
\toprule
\multirow{3}{*}{\textbf{Conflict Type}}
& \multicolumn{4}{c}{\textbf{R-LiViT}}
& \multicolumn{2}{c}{\textbf{Field}} \\
\cmidrule(lr){2-5} \cmidrule(lr){6-7}
& \multicolumn{2}{c}{\textbf{Int. 2}}
& \multicolumn{2}{c}{\textbf{Int. 3}}
& \multicolumn{2}{c}{\textbf{Georgia \& M.L.K.}} \\
\cmidrule(lr){2-3} \cmidrule(lr){4-5} \cmidrule(lr){6-7}
& \textbf{Pairs} & \textbf{Conflicts}
& \textbf{Pairs} & \textbf{Conflicts}
& \textbf{Pairs} & \textbf{Conflicts} \\
\midrule
Vehicle--Vehicle (TTC)  & 284 & 597  & 307 & 800  & 257 & 1190 \\
\midrule
Vehicle--Vehicle (PPET) & 218 & 287  &  96 & 130  & 783 & 1139 \\
Vehicle--VRU (PPET)     & 168 & 243  & 175 & 241  &   6 &    6 \\
VRU--VRU (PPET)         &  78 & 101  & 212 & 403  &   0 &    0 \\
\bottomrule
\end{tabularx}
\end{table}

In contrast, the Field site records only six Vehicle–VRU pairs and no VRU–VRU pairs, despite 106 VRU tracks in evaluation. The low Vehicle--VRU conflict count reflects dedicated pedestrian signal phases that temporally separate conflicting flows, whereas the absence of VRU--VRU conflicts results from cyclists and pedestrians traveling in distinct streams at the Field site, removing the primary source of co-moving agent pairs observed at R-LiViT. The Field site nonetheless exhibits a higher Vehicle--Vehicle (PPET) count (783 pairs) compared to R-LiViT (Int.2: 218 and Int.3: 96). This is partially attributable to a straight-motion bias in the prediction model, whereby decelerating or turning vehicles are initially predicted to continue straight, generating false cross-traffic conflicts in the early frames of a maneuver.

\subsection{Computational Performance and Real-Time Operation}

We evaluate computational performance on the NVIDIA Jetson AGX Thor under two LiDAR perception modes. In the deployed configuration, LiDAR perception is provided by the Ouster BlueCity API at 10\,Hz. We additionally benchmark a fully on-device configuration that replaces BlueCity with PointPillars~\cite{lang2019pointpillars} using the CUDA-accelerated TensorRT (FP16) implementation from the NVIDIA LiDAR AI Solution~\cite{nvidia_lidar_ai_solution} and AB3DMOT~\cite{weng2020ab3dmot} for multi-object tracking. Table~\ref{tab:runtime} presents the per-frame runtime breakdown of the fully on-device pipeline for an average scene density of 20--30 objects per frame. The reported trajectory prediction time includes data preparation (environment construction and DataLoader initialization), whereas Table~\ref{tab:traj_pred_all} reports model inference time only.

\begin{table}[h]
\captionsetup{justification=justified, singlelinecheck=false}
\centering
\small
\caption{Per-Frame Processing Time on Jetson AGX Thor}
\label{tab:runtime}
\begin{tabular}{lc}
\hline
\textbf{Module} & \textbf{Time (ms)} \\
\hline
Perception (PointPillars + AB3DMOT)          & 21  \\
Trajectory Prediction                        & 143 \\
SSA                                          & 30  \\
\hline
\textbf{Total}                               & \textbf{194} \\
\hline
\end{tabular}
\end{table}

Specifically, perception, trajectory prediction, and SSA contribute 21\,ms, 143\,ms, and 30\,ms, respectively, yielding a total end-to-end latency of $\delta = 194$\,ms. A risk assessment derived from observations at time $t$ therefore becomes available at $t + \delta$. Since FlowChain predicts trajectories over a 2.4\,s horizon, the usable prediction window for downstream decision-making is $2.4 - \delta \approx 2.21$\,s. Of the two safety metrics, only PPET relies on predicted trajectories, while TTC is computed directly from observed positions and velocities and is therefore unaffected by prediction latency.

%% file: sec/ablation.tex
\section{Ablation Study}

To validate the system's ability to operate without manual intervention, we conduct an ablation study evaluating the effectiveness of the proposed automated data processing module. Table~\ref{tab:data_processing} compares trajectory prediction accuracy between ground-truth annotations and the proposed data processing pipeline operating on raw perception outputs, evaluated on 4{,}000 field data frames at 5\,Hz under the same experimental configuration. The ADE and FDE gaps are 24.1\,\% and 26.2\,\%, respectively, demonstrating that the data processing module produces inputs of sufficient quality for reliable trajectory prediction without requiring any manual annotation.

\begin{table}[h]
\captionsetup{justification=justified, singlelinecheck=false}
\centering
\setlength{\tabcolsep}{4pt}
\renewcommand{\arraystretch}{1.1}
\small
\caption{Trajectory prediction accuracy of the proposed data processing pipeline versus ground-truth annotations.}
\label{tab:data_processing}
\begin{tabularx}{\columnwidth}{@{} l
    >{\centering\arraybackslash}X
    >{\centering\arraybackslash}X @{}}
\toprule
\textbf{Input} & \textbf{ADE (m)} & \textbf{FDE (m)} \\
\midrule
Ground-Truth Annotations  & 0.194 & 0.272 \\
Raw Perception + Data Processing     & 0.255 & 0.369 \\
\midrule
Gap (\%)                  & 24.1  & 26.2  \\
\bottomrule
\end{tabularx}
\end{table}

%% file: sec/conclusion.tex
\section{Conclusion}
We presented PRISA, a modular infrastructure LiDAR framework for proactive intersection safety assessment, integrating a self-supervised data curation pipeline, a learning-based trajectory prediction model, and a dual-SSM risk assessment module into a cohesive end-to-end system. By decoupling the plug-and-play risk assessment module from the upstream perception layer, PRISA is compatible with any compliant detection and tracking system, enabling deployment across heterogeneous roadside environments without perception-specific adaptation.

Experiments on the R-LiViT dataset and a live field deployment at the Georgia Avenue and M.L.K. Boulevard intersection in Chattanooga, Tennessee, demonstrate the practical viability of the proposed approach. The dual-SSM strategy — applying TTC for longitudinal conflicts and PPET for crossing and VRU-involved interactions — provides broad conflict coverage with configurable thresholds across diverse intersection geometries. While the end-to-end pipeline operates at 194 ms latency on an NVIDIA Jetson AGX Thor, this delay affects only PPET-based assessment; TTC and perception remain within real-time constraints, confirming the system's feasibility for continuous edge deployment.

Several directions remain open for future work. Incorporating maneuver-aware or intent-conditioned prediction models could reduce the straight-motion bias observed at the field site, while adding interaction-intent classification could help distinguish genuinely safety-critical encounters from co-moving agents in VRU-dense environments. Integrating lane geometry into the interaction classifier could further suppress false crossing detections and improve robustness. In addition, more detailed trajectory-level analysis under dense traffic flow and partial occlusion conditions would help characterize failure cases more systematically and guide further improvements in prediction robustness. Beyond LiDAR, the modular design could naturally accommodate camera-based or sensor-fusion perception systems, opening a pathway toward more robust multi-modal intersection safety monitoring.